%% file: iclr2027_conference_arxiv.tex
\documentclass[11pt]{article}

\usepackage[margin=1in]{geometry}
\usepackage{times}
\usepackage{url}
\usepackage{graphicx}
\usepackage{wrapfig}
\usepackage{placeins}
\usepackage{booktabs}
\usepackage{multirow}
\usepackage{xcolor}
\usepackage{makecell}
\usepackage{amsmath}
\usepackage{natbib}
\usepackage{caption}
\usepackage[hidelinks]{hyperref}

\input{math_commands.tex}

\definecolor{deepgreen}{RGB}{0,100,50}
\newcommand{\res}[2]{#1$\times$ & #2}
\newcommand{\bres}[2]{\textbf{#1$\times$} & \textbf{#2}}

\title{D3D: Decoupled Diffusion Drafting for Asynchronous Speculative Decoding}

\author{%
  Baoyou Chen\textsuperscript{1,2,3},
  Ruiqiao Mei\textsuperscript{2,3},
  Guancong Lin\textsuperscript{2,3},
  Tao Song\textsuperscript{2}\\[0.3em]
  Weihao Yuan\textsuperscript{4},
  Qun Liu\textsuperscript{5},
  Siyu Zhu\textsuperscript{1,2,3}\\[0.9em]
  {
  \begin{tabular}{@{}c@{\hspace{1.5em}}c@{}}
    \textsuperscript{1} Shanghai Innovation Institute &
    \textsuperscript{2} Shanghai Academy of AI for Science
  \end{tabular}}\\[0.3em]
  {
  \begin{tabular}{@{}c@{\hspace{1.5em}}c@{\hspace{1.5em}}c@{}}
    \textsuperscript{3} Fudan University &
    \textsuperscript{4} Nanjing University &
    \textsuperscript{5} Huawei Technologies Co., Ltd.
  \end{tabular}}
}
\date{}

\begin{document}
\maketitle

\begin{abstract}
Target-conditioned diffusion drafters predict candidate tokens in parallel, yet each new block must wait for updated target-model hidden states. This forces drafting and verification to run sequentially, limiting speedup even though each block is generated in one pass. We introduce D3D, a standalone diffusion drafter that removes this inter-round dependency. D3D combines blockwise denoising with auxiliary autoregressive supervision to achieve long accepted prefixes without target hidden states. This independence lets D3D draft successor blocks during verification in one packed pass, enabling asynchronous speculative decoding. We progressively prune layers according to their effect on acceptance length to reduce the cost of packed drafting. Removing five of 36 layers retains over 99\% of mean acceptance length across three benchmarks with Qwen3-8B as the target. Across seven benchmarks, D3D averages 5.00$\times$ and 5.71$\times$ decoding speedup with Qwen3-8B and Qwen3-32B targets, compared with 4.03$\times$ and 3.67$\times$ for DFlash. On GSM8K with Qwen3-32B, speedup reaches 7.48$\times$. Code is publicly available at \url{https://github.com/fudan-generative-vision/D3D}.
\end{abstract}

\input{sections/introduction}
\input{sections/relatedwork}
\input{sections/method}
\input{sections/experiments}
\input{sections/conclusion}



\bibliographystyle{plainnat}
\bibliography{iclr2027_conference}

\clearpage
\appendix
\input{sections/appendix}

\end{document}

%% file: math_commands.tex
\usepackage{amsmath,amsfonts,bm}

\def\eqref#1{equation~\ref{#1}}

\def\1{\bm{1}}

\DeclareMathAlphabet{\mathsfit}{\encodingdefault}{\sfdefault}{m}{sl}
\SetMathAlphabet{\mathsfit}{bold}{\encodingdefault}{\sfdefault}{bx}{n}



%% file: sections/introduction.tex
\section{Introduction}

Autoregressive language models dominate open-ended text generation, but their inherently sequential decoding limits token-level parallelism. Generating one token at a time leads to poor hardware utilization and high inference latency, especially in memory-bound settings. Speculative decoding alleviates this bottleneck by using a lightweight draft model to propose multiple future tokens, which the target model then verifies in parallel~\citep{sternBlockwiseParallelDecoding2018,xiaSpeculativeDecodingExploiting2023,leviathanFastInferenceTransformers2023a,chenAcceleratingLargeLanguage2023,kimSpeculativeDecodingBig2023,miaoSpecInferAcceleratingLarge2024}. With a suitable acceptance rule, speculative decoding preserves the target distribution, yielding the same output distribution as standard decoding with the target model.

The speedup achieved by speculative decoding depends on the number of accepted tokens, draft latency, and verification cost. Recent work has therefore focused on improving draft accuracy while keeping draft latency low. Many methods attach auxiliary predictors to the target model or condition a separate drafter on intermediate representations from the target model~\citep{caiMedusaSimpleLLM2024,anknerHydraSequentiallyDependentDraft2024,gloeckleBetterFasterLarge2024,liEAGLESpeculativeSampling2025a,liEagle3ScalingInference2026,chengRecurrentDrafterFast2024,xiaoCloverRegressiveLightweight2024}. However, autoregressive and recurrent drafters still generate candidate tokens sequentially, causing draft latency to grow with the candidate length.

Parallel drafting addresses the sequential generation cost by predicting multiple candidate positions simultaneously. PARD adapts a pretrained autoregressive model into a parallel drafter, while DiffuSpec and SpecDiff-2 use diffusion models to generate multiple candidate tokens concurrently~\citep{anPARDAcceleratingLLM2025,liDiffuspecUnlockingDiffusion2026,sandlerSpecDiff2ScalingDiffusion2025}. PARD-2 further aligns training with acceptance length and supports both target-dependent and target-independent inference~\citep{an2026pard2}. DART and DFlash improve draft quality by conditioning parallel drafters on intermediate representations from the target model~\citep{liuDARTDiffusionInspiredSpeculative2026,chenDFlashBlockDiffusion2026a}. In particular, DFlash uses a block diffusion model to generate an entire candidate block in a single forward pass. Recent methods improve target-conditioned parallel drafts by enriching target-feature fusion \citep{zhangDFlareScalingDraft2026}, refining block predictions with causal context \citep{huang2026domino,rheinboldt2026treeflash,zheng2026delsspec}, and weighting training positions according to their contribution to acceptance \citep{wu2026dpace}. Target-conditioned parallel heads have also been used to generate causal draft trees \citep{hu2026jetspec}. In these implementations, the next draft block requires updated target-model representations.

For a diffusion drafter that generates an entire candidate block in a single forward pass, drafting latency still depends on the drafter's depth. Recent work reports a log-linear relationship between decoder depth and acceptance rate for shallow autoregressive drafters~\citep{yanScalingLawsSpeculative2025}. Larger drafters can improve acceptance, but typically require more time to generate each candidate block. When drafting and verification execute sequentially, this additional draft latency can offset the benefit of accepting more tokens. This creates a trade-off between draft quality and draft latency.


Beyond the cost of each forward pass, conditioning on fresh target-model representations creates an execution dependency between drafting rounds. In DFlash, verification produces representations of newly accepted tokens that the drafter consumes in the next round~\citep{chenDFlashBlockDiffusion2026a}. The next draft must therefore wait for these representations, even though the drafter predicts all positions within a block in parallel.


Inspired by Speculative Speculative Decoding (SSD), we prepare drafts for possible verification outcomes while the target verifies the current block~\citep{kumarSpeculativeSpeculativeDecoding2026}. SSD demonstrates this approach with autoregressive drafters that generate candidate sequences through multiple sequential steps. We use a standalone diffusion drafter to predict each candidate block from its conditioning prefix in a single forward pass. The modeling challenge is to obtain long accepted prefixes without target hidden states while keeping drafting fast enough to complete within the verification interval.

We introduce D3D, a decoupled diffusion drafter built around target-aligned training and progressive layer pruning. Our contributions are as follows:
\begin{enumerate}
\item \textbf{Target-aligned standalone diffusion drafting.} We combine blockwise denoising and auxiliary autoregressive supervision on target-generated responses to train a standalone, single-pass drafter from pretrained weights. At block size $16$, D3D exceeds every baseline in acceptance length on each of the seven benchmarks for both targets, averaging $7.88$ with Qwen3-8B and $7.73$ with Qwen3-32B (Table~\ref{tab:dflash_speedup}).

\item \textbf{Pruning redundant drafter layers.} We remove layers with the smallest effect on validation acceptance length, with recovery training between removals. With Qwen3-8B as the target, removing five of $36$ layers ($13.9\%$) retains over $99\%$ of mean acceptance length and raises mean speedup from $4.44\times$ to $4.69\times$ across three benchmarks (Figure~\ref{fig:layer_pruning_ablation}).

\item \textbf{Asynchronous speculative decoding.} During target verification, D3D prepares possible successor blocks in one packed drafter pass and reuses a matching block in the next round. Across seven benchmarks, it achieves the highest mean decoding speedup over autoregressive decoding among the evaluated methods: $5.00\times$ with Qwen3-8B and $5.71\times$ with Qwen3-32B (Table~\ref{tab:dflash_speedup}).
\end{enumerate}

%% file: sections/relatedwork.tex
\section{Related Work}

\subsection{Parallel and Diffusion Drafting}
PARD adapts an autoregressive backbone for masked parallel prediction~\citep{anPARDAcceleratingLLM2025}. PARD-2 uses causal masked-token prediction and stochastically injects target features during training, allowing the same drafter to operate with or without these features at inference~\citep{an2026pard2}. DFlash injects projected target features into the attention keys and values of each diffusion layer~\citep{chenDFlashBlockDiffusion2026a}. D$^2$SD retains this conditioning in both its initial drafter and its second drafter, which generates continuations from variable-length prefixes~\citep{zhang2026d2sd}. D$^2$SD jointly verifies these alternative continuations in the current round. D3D prepares successor blocks for possible next-round prefixes while the target verifies the current block. DeLS-Spec independently trains a short-context predictor and combines its logits with those of a frozen DFlash backbone~\citep{zheng2026delsspec}. Its decoupling separates component training, while the backbone continues to consume target features. D3D shares target-independent inference with PARD and PARD-2, and learns its own prefix representations through diffusion and autoregressive supervision on target-generated responses. Its masked blocks use bidirectional attention, and training does not inject target hidden states. For D3D, we further prune layers based on their measured effect on acceptance length.

SpecDiff-2 derives streak distillation from a surrogate for the expected accepted prefix length~\citep{sandlerSpecDiff2ScalingDiffusion2025}. PARD-2 weights token losses using the cumulative product of the target model's confidence in preceding tokens~\citep{an2026pard2}. D-PACE derives example-dependent position weights from a surrogate for expected accepted length~\citep{wu2026dpace}. DFlash uses exponentially decaying position weights to emphasize early predictions~\citep{chenDFlashBlockDiffusion2026a}. We retain DFlash's position weighting and pair blockwise denoising with auxiliary autoregressive supervision on the same target-generated responses. The auxiliary objective trains next-token prediction from unmasked response prefixes, adding a distinct conditioning pattern to the masked-block training objective.

Other methods optimize how candidate tokens are organized and selected for verification. DiffuSpec searches for causally consistent paths through parallel token predictions and adapts draft length using recent acceptance feedback~\citep{liDiffuspecUnlockingDiffusion2026}. SpecDiff-2 selects among multiple sampled drafts using a verifier-based criterion~\citep{sandlerSpecDiff2ScalingDiffusion2025}. DART and DDTree organize parallel predictions into candidate trees~\citep{liuDARTDiffusionInspiredSpeculative2026,ringelAcceleratingSpeculativeDecoding2026a}, while CaDDTree jointly selects tree structure and node budget using a throughput objective~\citep{zhangCostAwareDiffusionDraft2026}. DSpark couples a parallel backbone with a lightweight sequential module and selects verification length using prefix acceptance estimates and serving load~\citep{chengDSparkConfidenceScheduledSpeculative2026}. These methods modify candidate dependencies or verification choices; D3D studies the training and layer redundancy of the underlying standalone diffusion drafter.

\subsection{Layer Pruning for Efficient Drafting}
ShortGPT ranks layers by the similarity between their input and output representations~\citep{menShortGPTLayers2025}. SDFP uses Fisher information to select layers for removal and constructs drafters without recovery training~\citep{weiSDFPSpeculativeDecoding2026}. \citet{yanDecodingSpeculativeDecoding2025} study structured pruning for shallow draft architectures and show that language-model accuracy is a poor predictor of speculative-decoding performance. D3D evaluates layer removal through its effect on acceptance length, directly measuring the retained model's utility for block drafting.

Self-data distillation restores quality after layer pruning using responses generated by the unpruned model~\citep{thangarasaSelfDataDistillation2025}. SD$^2$ applies self-distillation to fine-grained sparse drafters and also evaluates layer-pruned alternatives~\citep{lasbySelfDistilledSparseDrafters2025}. Our procedure alternates layer removal with recovery training and reassesses acceptance-based layer importance after recovery, so each pruning decision reflects the updated drafter. Within diffusion drafting, DFlare studies greater depth with layer-specific fusion of target features~\citep{zhangDFlareScalingDraft2026}; D3D studies depth reduction while retaining nearly all of a trained standalone drafter's acceptance length.

\subsection{Asynchronous Speculative Decoding}
PipeInfer maintains speculative execution during pipeline-parallel inference and cancels computation along invalid paths~\citep{butlerPipeinferAcceleratingLlm2024}. SwiftSpec separates drafting and verification across GPU groups and coordinates parallel tree generation with cache management~\citep{zhang2025swiftspecultralowlatencyllm}. Speculative Speculative Decoding (SSD) caches follow-up drafts for possible verification outcomes and uses a fallback strategy when the observed outcome has no cached draft~\citep{kumarSpeculativeSpeculativeDecoding2026}. SSD explicitly accommodates different drafter architectures, including diffusion models. Inspired by SSD, D3D combines target-aligned drafter training and acceptance-guided pruning with packed successor prediction during target verification.


%% file: sections/method.tex
\section{Methodology}
\label{sec:method}

\begin{figure*}[t]
    \centering
    \includegraphics[width=0.95\textwidth]{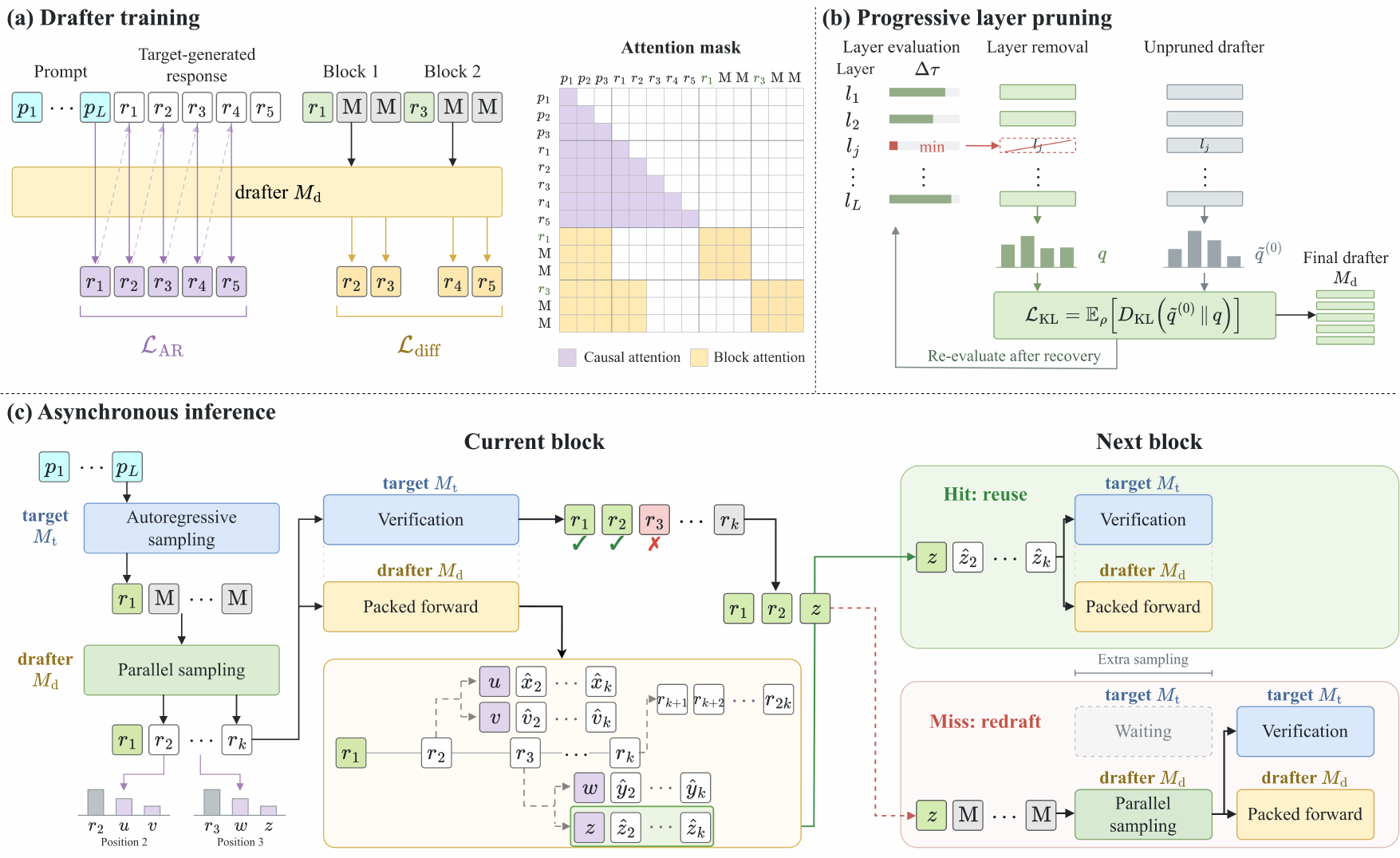}
\caption{Overview of D3D. \textbf{(a) Drafter training.} Target-generated responses supervise blockwise denoising and auxiliary autoregressive prediction. Packed attention shares causal prefix computation while isolating diffusion blocks and allowing bidirectional attention within each block. \textbf{(b) Progressive layer pruning.} Each iteration removes the layer causing the smallest decrease in validation acceptance length, followed by recovery and reevaluation of layer importance. Recovery uses distillation from a fixed unpruned drafter whose output distribution is adjusted toward the supervised token. \textbf{(c) Asynchronous inference.} During target verification, one packed drafter pass prepares successor blocks for alternative rejection outcomes and full acceptance. A branch matching the verified prefix is reused; otherwise, the drafter generates a new block synchronously.}
\label{fig:d3d_overview}
\end{figure*}

\subsection{Overview}
\label{sec:overview}
D3D is a standalone diffusion drafter $M_{\mathrm{d}}$ for asynchronous speculative decoding with an autoregressive target model $M_{\mathrm{t}}$. Conditioned on the token prefix, $M_{\mathrm{d}}$ computes its own prefix representations and generates a candidate block in a single forward pass, without accessing target hidden states. The target verifies the candidate tokens in parallel using rejection sampling and appends one additional token after the accepted prefix.

Our objective is to fully hide drafting latency within target verification while maintaining long accepted prefixes. When drafting is fully hidden, target verification determines the ideal latency of each decoding round. We initialize $M_{\mathrm{d}}$ from a pretrained autoregressive model and train on target-generated responses using blockwise denoising and auxiliary autoregressive supervision (Section~\ref{sec:target_aligned_drafter_training}). 
We then exploit redundancy across drafter layers through acceptance-guided pruning to reduce drafting latency while preserving the trained drafter’s acceptance length (Section~\ref{sec:acceptance_guided_layer_pruning}). Section~\ref{sec:inference} describes concurrent drafting and verification and the conditions for hiding drafting latency. Figure~\ref{fig:d3d_overview} summarizes the training, pruning, and inference stages.

\subsection{Target-Aligned Drafter Training}
\label{sec:target_aligned_drafter_training}

A standalone drafter must encode the token prefix and predict continuations using its own representations.
We initialize $M_{\mathrm{d}}$ from a pretrained autoregressive model to use its learned language priors. For each prompt $p=(p_1,\ldots,p_L)$, we generate a response $r=(r_1,\ldots,r_T)$ using $M_{\mathrm{t}}$ and train $M_{\mathrm{d}}$ on these response tokens.

We randomly sample $K$ anchor positions
\begin{equation}
\mathcal{S}(r)=\{s_1,\ldots,s_K\},\qquad 1\leq s_1<\cdots<s_K<T,
\end{equation}
and construct a diffusion block of size $B$ for each anchor:
\begin{equation}
\widetilde{B}_k=\bigl(r_{s_k},\underbrace{[\mathrm{M}],\ldots,[\mathrm{M}]}_{B-1}\bigr).
\end{equation}
Here, $[\mathrm{M}]$ denotes a mask token marking a position to be predicted. We use a reserved token from the pretrained model's vocabulary as $[\mathrm{M}]$. The anchor token $r_{s_k}$ remains visible, and the drafter predicts the $B-1$ masked positions in parallel. For each block, only the next $b_k=\min(B-1,T-s_k)$ response tokens are supervised. Masked positions beyond the response boundary remain in the block and participate in attention but are excluded from the loss.


We pack the $K$ diffusion blocks into a single sequence to obtain denoising supervision for the masked positions following each anchor in one forward pass (Figure~\ref{fig:d3d_overview}(a)). $M_{\mathrm{d}}$ computes a shared causal encoding of $(p,r)$ to avoid redundant prefix computation. Each block $\tilde{B}_k$ attends only to the representations of
$(p, r_{<s_k})$ and its own tokens, with bidirectional attention
within the block.

For the $j$-th masked position in the $k$-th block, we define
\begin{equation}
\ell^{\mathrm{diff}}_{k,j}=-\log P_{M_{\mathrm{d}}}\!\left(r_{s_k+j}\mid p,r_{<s_k},\tilde{B}_k\right).
\end{equation}
Here, $P_{M_{\mathrm{d}}}$ denotes the drafter's conditional token distribution.
Verification retains candidate tokens up to the first rejection, motivating greater weight on early block positions.
Following DFlash~\citep{chenDFlashBlockDiffusion2026a}, we use exponentially decaying position weights:
\begin{equation}
w_j=\exp\!\left(-\frac{j-1}{\gamma}\right),
\quad \text{where} \ j\in\{1,\ldots,B-1\}.
\end{equation}
The parameter $\gamma>0$ controls the decay rate.
The position-weighted diffusion objective is
\begin{equation}
\mathcal{L}_{\mathrm{diff}}
=\frac{1}{Z_{\mathrm{diff}}}
\sum_{k=1}^{K}\sum_{j=1}^{b_k}w_j\ell^{\mathrm{diff}}_{k,j},
\quad \text{where}\quad
Z_{\mathrm{diff}}
=\sum_{k=1}^{K}\sum_{j=1}^{b_k}w_j.
\end{equation}

To preserve the pretrained model's language modeling capability, we apply an auxiliary autoregressive objective to the unmasked target-generated response:
\begin{equation}
\mathcal{L}_{\mathrm{AR}}=-\frac{1}{T}\sum_{i=1}^{T}\log P_{M_{\mathrm{d}}}(r_i\mid p,r_{<i}).
\end{equation}
The complete training objective is
\begin{equation}
\mathcal{L}_{\mathrm{train}}= \mathcal{L}_{\mathrm{diff}} + \lambda\mathcal{L}_{\mathrm{AR}},
\label{eq:drafter_training}
\end{equation}
where $\lambda$ controls the contribution of the autoregressive objective.

\subsection{Acceptance-Guided Layer Pruning}
\label{sec:acceptance_guided_layer_pruning}

We exploit redundancy across drafter layers to reduce drafting latency while preserving average acceptance length. With the target model and block size fixed, we measure validation acceptance length after separately removing each remaining layer (Figure~\ref{fig:d3d_overview}(b)).
We remove the single layer whose removal causes the smallest decrease in average acceptance length, then perform recovery training. Layer removal changes the inputs to subsequent layers, and recovery training updates the remaining parameters. We therefore recompute layer importance on the recovered model before the next pruning step.

Recovery training uses knowledge distillation~\citep{hintonDistillingKnowledgeNeural2015} from the trained, unpruned drafter, which remains fixed throughout pruning.
At each supervised prediction position, let $y$ denote the supervised token from the target-generated response, and let $q^{(0)}$ and $q$ denote the teacher and pruned drafter distributions.
We compute both distributions with the same block input and attention pattern.
Direct distillation can favor a competing token when the teacher's most probable prediction differs from $y$. We adjust the teacher distribution toward the supervised token $y$ by mixing it with the one-hot distribution $e_y$:
\begin{equation}
\tilde q^{(0)}=(1-\eta)q^{(0)}+\eta e_y.
\end{equation}
At each prediction position, we choose the smallest mixing weight $\eta$ that makes $y$ more probable than every other token by at least a fixed positive margin.
Choosing the smallest weight retains as much of the teacher's soft distribution as possible.
Appendix~\ref{app:teacher_distribution_adjustment} provides the closed-form expression for $\eta$.

The distillation loss is the position-weighted forward KL divergence from the adjusted teacher distribution to the pruned drafter:
\begin{equation}
\mathcal{L}_{\mathrm{KL}}
=\mathbb{E}_{\rho}\!\left[D_{\mathrm{KL}}\!\left(\tilde q^{(0)}\Vert q\right)\right].
\end{equation}
Here, $\mathbb{E}_{\rho}$ denotes a normalized average over masked prediction positions, with weights that decay exponentially with distance from each block's anchor.
We combine this loss with the training objective from Section~\ref{sec:target_aligned_drafter_training}:
\begin{equation}
\mathcal{L}_{\mathrm{rec}}
=\mathcal{L}_{\mathrm{diff}}+\lambda\mathcal{L}_{\mathrm{AR}}+\mathcal{L}_{\mathrm{KL}}.
\end{equation}


\subsection{Inference}
\label{sec:inference}
$M_{\mathrm{t}}$ generates the first response token as the initial anchor, and $M_{\mathrm{d}}$ predicts $B-1$ candidate tokens in a single forward pass, forming a block of size $B$.
The target verifies these candidates in parallel using rejection sampling with their corresponding proposal probabilities~\citep{leviathanFastInferenceTransformers2023a}.
It appends the accepted candidates and samples the next token from the residual distribution at the first rejection, or from the target distribution if all candidates are accepted.

During verification, $M_{\mathrm{d}}$ prepares continuations for possible verification outcomes in a single packed forward pass (Figure~\ref{fig:d3d_overview}(c)).
Each rejection branch extends the existing output prefix with the candidates preceding the rejection position, an alternative anchor, and a predicted continuation.
For full acceptance, the drafter also prepares a continuation conditioned on the entire current block.

A cache hit occurs when the complete verified prefix, including the target-generated token, matches the initial token sequence of a prepared branch.
We reuse the remaining draft tokens of the matching branch for the next verification round.
On a cache miss, the target must wait for $M_{\mathrm{d}}$ to synchronously generate a new block from the verified prefix before the next verification round can begin.
We define the cache hit rate $p_{\mathrm{hit}}$ as the probability of a hit in a verification round.

Position coverage $C$ includes the first $C-1$ rejection positions and the full-acceptance case.
Token fan-out $F$ is the number of alternative anchors per selected rejection position~\citep{kumarSpeculativeSpeculativeDecoding2026}.
We select the $F$ most probable tokens under the drafter distribution, excluding the proposed token; under greedy drafting, these have ranks $2,\ldots,F+1$.

The packed input contains the current block which is being verified, $(C-1)F$ blocks for rejection outcomes, and one block for full acceptance.
With $B$ token positions per block, the asynchronous draft budget is
\begin{equation}
N_{\mathrm{async}}=B\left[(C-1)F+2\right].
\label{eq:async_draft_budget}
\end{equation}

Let $t_{\mathrm{d}}$ and $t_{\mathrm{v}}$ denote the latencies of packed drafting and target verification under concurrent execution.
For stages that begin simultaneously, a cache hit hides the packed drafting latency when $t_{\mathrm{d}}\leq t_{\mathrm{v}}$.
Larger budgets can increase $t_{\mathrm{d}}$ and reduce the latency benefit obtained from a higher cache-hit rate.
Acceptance-guided pruning aims to reduce $t_{\mathrm{d}}$ with minimal loss in acceptance length.

%% file: sections/experiments.tex
\section{Experiments}

\subsection{Experimental Setup}

\paragraph{Models and Evaluation}
We evaluate D3D with Qwen3-8B and Qwen3-32B as target models across seven benchmarks.
These cover mathematical reasoning (GSM8K, MATH-500, and AIME25), code generation (HumanEval, MBPP, and LiveCodeBench-v6), and open-ended dialogue (MT-Bench).
We report decoding speedup relative to autoregressive decoding and average acceptance length $\tau$.
We define $\tau$ as the mean number of output tokens added per verification round, including the token generated by the target model.

\paragraph{Training Data}
We use prompts from Open-PerfectBlend, an instruction-tuning dataset containing 1.42 million examples. We regenerate responses with the corresponding target model in non-thinking mode, using its recommended sampling settings.

\paragraph{Baselines}
We compare D3D with DFlash~\citep{chenDFlashBlockDiffusion2026a},
DSpark~\citep{chengDSparkConfidenceScheduledSpeculative2026},
EAGLE-3~\citep{liEagle3ScalingInference2026},
PARD-2~\citep{an2026pard2}, and SSD~\citep{kumarSpeculativeSpeculativeDecoding2026}.
PARD-2 and SSD are evaluated using asynchronous inference.
For Qwen3-8B, we use the public DSpark and EAGLE-3 checkpoints released by the DSpark authors, both trained on Open-PerfectBlend.
We train DFlash and PARD-2 for both targets, and DSpark and EAGLE-3 for Qwen3-32B, on the same target-generated training data as D3D.

\paragraph{Implementation}
The main experiments use D3D-8B, initialized from Qwen3-8B and progressively pruned from 36 to 31 Transformer layers, with $B=16$, $C=12$, and $F=2$ for both targets (Section~\ref{sec:inference}). Target-aligned drafter training uses 128 random anchors per example and 3 training epochs (360 GPU hours). Target-aligned drafter training and recovery training use BF16 on two NVIDIA H800 GPUs with 8-bit AdamW.
We use a local batch size of 1 and gradient accumulation steps of 8, giving a global batch size of 16.
We use a cosine learning-rate schedule from $10^{-5}$ to $10^{-6}$ with 1\% warmup, zero weight decay, and gradient-norm clipping at 1.

For pruning, we reuse the target-aligned training data and perform 10k recovery training steps per stage. Layer selection and checkpoint selection are performed on two separate, non-overlapping development sets, with acceptance length evaluated on 512 examples per set.

All evaluations use the Transformers backend with greedy decoding in non-thinking mode. D3D, SSD, and PARD-2 use two H800 GPUs; the synchronous DFlash, DSpark, and EAGLE-3 baselines run on one.

\subsection{Main Results}
\begin{table}[!ht]
    \centering
    \setlength{\tabcolsep}{2.2pt}
    \renewcommand{\arraystretch}{1.12}
    \caption{Speedup and average acceptance length $\tau$ across mathematical
    reasoning, code generation, and chat benchmarks. LCB denotes LiveCodeBench. Numbers in parentheses indicate block sizes.}
    \resizebox{\textwidth}{!}{%
    \begin{tabular}{@{}ll*{8}{cc}@{}}
        \toprule

        \multirow{2}{*}{Model}
        & \multirow{2}{*}{Method}
        & \multicolumn{6}{c}{\textsc{Math}}
        & \multicolumn{6}{c}{\textsc{Code}}
        & \multicolumn{2}{c}{\textsc{Chat}}
        & \multicolumn{2}{c}{} \\
        \cmidrule(lr){3-8}
        \cmidrule(lr){9-14}
        \cmidrule(lr){15-16}

        &
        & \multicolumn{2}{c}{GSM8K}
        & \multicolumn{2}{c}{MATH-500}
        & \multicolumn{2}{c}{AIME25}
        & \multicolumn{2}{c}{HumanEval}
        & \multicolumn{2}{c}{MBPP}
        & \multicolumn{2}{c}{LCB-v6}
        & \multicolumn{2}{c}{MT-Bench}
        & \multicolumn{2}{c}{\emph{Avg.}} \\
        \midrule

        \multicolumn{2}{c}{}
        & Speedup & $\tau$
        & Speedup & $\tau$
        & Speedup & $\tau$
        & Speedup & $\tau$
        & Speedup & $\tau$
        & Speedup & $\tau$
        & Speedup & $\tau$
        & Speedup & $\tau$ \\
        \midrule

        \multirow{6}{*}{\shortstack[c]{Qwen3\\8B}}

        & EAGLE-3 \textcolor{gray}{(8)}
        & \res{3.43}{5.58}
        & \res{3.34}{5.38}
        & \res{3.01}{4.76}
        & \res{2.95}{4.71}
        & \res{2.71}{4.31}
        & \res{2.88}{4.30}
        & \res{1.87}{2.90}
        & \res{2.88}{4.56} \\


        & DFlash \textcolor{gray}{(16)}
        & \res{4.59}{7.93}
        & \res{5.00}{7.44}
        & \res{4.34}{6.29}
        & \res{4.34}{6.02}
        & \res{4.03}{5.29}
        & \res{3.48}{5.32}
        & \res{2.46}{3.48}
        & \res{4.03}{5.96} \\


        & DSpark \textcolor{gray}{(8)}
        & \res{4.93}{6.47}
        & \res{4.86}{6.40}
        & \res{4.79}{5.75}
        & \res{4.70}{5.84}
        & \res{4.19}{5.48}
        & \res{4.44}{5.40}
        & \res{3.17}{3.82}
        & \res{4.44}{5.59} \\

        & PARD-2 \textcolor{gray}{(16)}
        & \res{6.16}{7.26}
        & \res{5.81}{7.71}
        & \res{4.44}{5.52}
        & \res{3.95}{5.46}
        & \res{3.39}{4.78}
        & \res{2.89}{4.43}
        & \res{1.89}{4.15}
        & \res{4.08}{5.62} \\

        & SSD \textcolor{gray}{(4)}
        & \res{1.98}{3.43}
        & \res{2.06}{3.53}
        & \res{2.00}{3.44}
        & \res{1.87}{3.26}
        & \res{1.80}{3.11}
        & \res{1.83}{3.17}
        & \res{1.34}{2.79}
        & \res{1.84}{3.25} \\

        & D3D \textcolor{gray}{(16)}
        & \bres{6.71}{9.87}
        & \bres{6.21}{9.84}
        & \bres{5.11}{7.98}
        & \bres{5.19}{7.79}
        & \bres{4.70}{7.31}
        & \bres{4.44}{6.68}
        & \bres{2.68}{5.71}
        & \bres{5.00}{7.88} \\




        \midrule

        \multirow{6}{*}{\shortstack[c]{Qwen3\\32B}}
        & EAGLE-3 \textcolor{gray}{(8)}
        & \res{4.08}{6.05}
        & \res{4.16}{5.92}
        & \res{3.53}{5.36}
        & \res{3.72}{5.25}
        & \res{3.54}{4.96}
        & \res{2.91}{4.61}
        & \res{2.31}{3.44}
        & \res{3.46}{5.08} \\

        & DFlash \textcolor{gray}{(16)}
        & \res{4.46}{5.72}
        & \res{4.37}{5.57}
        & \res{3.99}{5.28}
        & \res{3.75}{4.71}
        & \res{3.48}{4.50}
        & \res{2.66}{4.04}
        & \res{3.00}{3.89}
        & \res{3.67}{4.82} \\

        & DSpark \textcolor{gray}{(8)}
        & \res{4.59}{6.41}
        & \res{4.40}{6.34}
        & \res{3.98}{5.79}
        & \res{4.25}{5.66}
        & \res{4.33}{5.39}
        & \res{3.58}{5.07}
        & \res{2.35}{3.66}
        & \res{3.92}{5.48} \\

        & PARD-2 \textcolor{gray}{(16)}
        & \res{5.62}{6.89}
        & \res{5.38}{7.64}
        & \res{4.02}{5.27}
        & \res{3.92}{5.14}
        & \res{3.24}{4.55}
        & \res{2.82}{4.33}
        & \res{2.19}{3.86}
        & \res{3.88}{5.38} \\

        & SSD \textcolor{gray}{(6)}
        & \res{3.84}{4.62}
        & \res{4.05}{4.77}
        & \res{3.97}{4.61}
        & \res{3.56}{4.11}
        & \res{3.32}{3.92}
        & \res{3.65}{4.14}
        & \res{2.35}{3.41}
        & \res{3.54}{4.22} \\
        





        & D3D \textcolor{gray}{(16)}
        & \bres{7.48}{9.58}
        & \bres{7.46}{9.68}
        & \bres{5.85}{8.18}
        & \bres{6.22}{7.66}
        & \bres{5.71}{7.20}
        & \bres{4.25}{6.13}
        & \bres{3.01}{5.65}
        & \bres{5.71}{7.73} \\

        \bottomrule
    \end{tabular}%
    }
    \label{tab:dflash_speedup}
\end{table}

\paragraph{Decoding performance.}
D3D achieves higher average decoding speedup than DFlash,
DSpark, EAGLE-3, PARD-2, and SSD for both target models
(Table~\ref{tab:dflash_speedup}).
With Qwen3-8B, D3D reaches an average speedup of $5.00\times$, compared with $4.44\times$ for DSpark and $4.03\times$ for DFlash.
With Qwen3-32B, D3D reaches $5.71\times$, compared with $3.92\times$ and $3.67\times$, respectively.
These correspond to improvements of $12.6\%$ and $45.7\%$ over the highest baseline average in our evaluation.
The peak speedup reaches $7.48\times$ on GSM8K with Qwen3-32B.

D3D also achieves longer accepted sequences than all five baselines on every benchmark for both targets.
Its average $\tau$ is $7.88$ with Qwen3-8B and $7.73$ with Qwen3-32B, compared with DFlash's $5.96$ and $4.82$ at the same block size.
These results show that conditioning on the token prefix alone supports high acceptance across both target sizes.
The speedup advantage varies by benchmark: D3D is fastest on five of seven benchmarks with Qwen3-8B and all seven with Qwen3-32B.

\paragraph{Concurrent serving.}
\begin{wraptable}[9]{r}{0.46\textwidth}
  \centering
  \caption{Decoding throughput in \mbox{tokens/s} (speedup in parentheses) versus concurrency; D3D-1.7B drafter, Qwen3-32B target.}
  \label{tab:concurrent_throughput}
  \footnotesize
  \setlength{\tabcolsep}{2pt}
  \begin{tabular}{@{}rccc@{}}
    \toprule
    Requests & MATH-500 & HumanEval & MT-Bench \\
    \midrule
     1 & $154\,(6.1\times)$ & $135\,(5.3\times)$ & $75\,(2.9\times)$ \\
     4 & $470\,(4.8\times)$ & $407\,(4.3\times)$ & $202\,(2.1\times)$ \\
     8 & $745\,(3.9\times)$ & $655\,(3.9\times)$ & $327\,(1.9\times)$ \\
    16 & $864\,(2.7\times)$ & $748\,(2.6\times)$ & $386\,(1.5\times)$ \\
    \bottomrule
  \end{tabular}
\end{wraptable}
Using the Transformers backend with SDPA, we measure aggregate decoding throughput with D3D-1.7B and Qwen3-32B across 1--16 concurrent requests (Table~\ref{tab:concurrent_throughput}). At 16 requests, MATH-500 reaches 864 tokens/s ($2.7\times$), while MT-Bench reaches 386 tokens/s ($1.5\times$). Throughput rises with concurrency on all three tasks, while relative speedup declines. Single-request speedup therefore does not capture how the gain changes under load.

\subsection{Ablation Studies}

\paragraph{Auxiliary autoregressive supervision.}
We assess the contribution of auxiliary AR supervision by varying $\lambda$ in Eq.~(\ref{eq:drafter_training}) and training each D3D-1.7B variant for one epoch (Table~\ref{tab:ar_loss_ablation}, left).
Setting $\lambda=1$ increases $\tau$ by $0.24$ to $0.40$ tokens over diffusion-only training ($\lambda=0$) across all four benchmarks; increasing $\lambda$ to $1.5$ provides no further improvement.
The GSM8K training curves show that the benefit persists throughout training (Appendix~\ref{app:training_dynamics}).
Blockwise denoising trains parallel predictions, while the AR objective continues to supervise next-token prediction from unmasked prefixes.
This supervision helps the standalone drafter learn to predict from its own prefix representations without target hidden states.
We therefore use $\lambda=1$ during target-aligned drafter training.

\paragraph{Drafter scale.}
The scale ablation examines whether D3D's acceptance advantage requires a larger parameter budget.
We train its 1.7B, 4B, and 8B variants for one epoch and compare them with DFlash (Table~\ref{tab:ar_loss_ablation}, right).
D3D-1.7B achieves higher acceptance than DFlash on all three benchmarks with fewer total parameters ($2.0$B versus $2.3$B).
On MATH-500, increasing drafter size from 1.7B to 4B improves $\tau$ by $1.29$ tokens, while increasing it from 4B to 8B adds $0.44$ tokens; HumanEval and MT-Bench show similar diminishing gains.
The AR and initialization ablations separately show that both training choices improve acceptance at a fixed architecture.
The smaller gains at larger scales motivate pruning the trained 8B drafter to reduce drafting cost while retaining its acceptance.

\begin{table}[!hb]
  \centering
\caption{Effects of auxiliary AR weight (a) and drafter scale (b) on average acceptance length $\tau$.
We evaluate D3D variants after one epoch of training, with Qwen3-8B as the target and $B=16$.}
  \label{tab:ar_loss_ablation}
  \begin{minipage}[t]{0.43\linewidth}
    \centering\footnotesize
    (a) Auxiliary AR weight\par\smallskip
    \setlength{\tabcolsep}{2.5pt}
    \renewcommand{\arraystretch}{0.88}
    \begin{tabular}{@{}ccccc@{}}
      \toprule
      $\lambda$ & GSM8K & MATH-500 & HumanEval & MBPP \\
      \midrule
      $0$   & 7.89 & 7.86 & 5.86 & 5.53 \\
      $0.5$ & 8.05 & 7.98 & 6.02 & 5.64 \\
      $1.0$ & \textbf{8.29} & \textbf{8.16} & \textbf{6.25} & \textbf{5.77} \\
      $1.5$ & 8.25 & 8.13 & 6.19 & 5.75 \\
      \bottomrule
    \end{tabular}
  \end{minipage}\hfill
  \begin{minipage}[t]{0.53\linewidth}
    \centering\footnotesize
    (b) Drafter scale\par\smallskip
    \setlength{\tabcolsep}{2pt}
    \renewcommand{\arraystretch}{0.88}
    \begin{tabular}{@{}lcccc@{}}
      \toprule
      Drafter & Params. & MATH-500 & HumanEval & MT-Bench \\
      \midrule
      DFlash & 2.3B & 6.87 & 5.42 & 3.23 \\
      D3D-1.7B & 2.0B & 8.16 & 6.25 & 4.57 \\
      D3D-4B & 4.4B & 9.45 & 7.42 & 5.38 \\
      D3D-8B & 8.2B & \textbf{9.89} & \textbf{7.86} & \textbf{5.73} \\
      \bottomrule
    \end{tabular}
  \end{minipage}
\end{table}

\begingroup
\setlength{\intextsep}{6pt}
\begin{figure}[!htb]
  \centering
  \begin{minipage}[t]{0.49\linewidth}
    \centering
    \includegraphics[width=\linewidth]{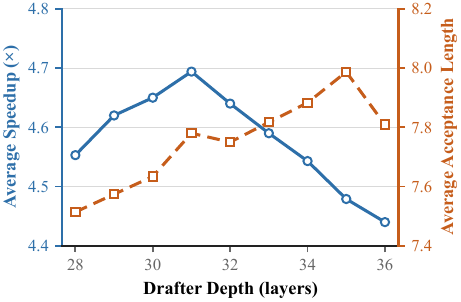}
  \end{minipage}
  \hspace{0.01\linewidth}%
  \begin{minipage}[t]{0.47\linewidth}
    \centering
    \includegraphics[width=\linewidth]{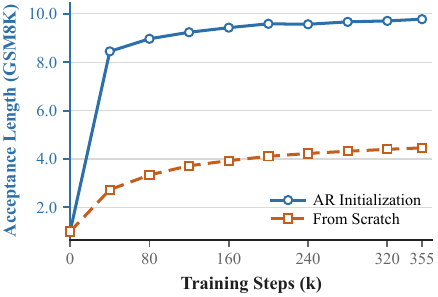}
  \end{minipage}
\caption{Ablations of D3D-8B depth and initialization with Qwen3-8B as the target and $B=16$. Left: decoding speedup and acceptance length $\tau$, averaged over MATH-500, HumanEval, and MT-Bench. Right: GSM8K $\tau$ during training from pretrained or random initialization.}
  \label{fig:layer_pruning_ablation}
\end{figure}
\endgroup

\paragraph{Acceptance-guided layer pruning.}
To test whether depth reduction improves speedup with little loss in acceptance, we evaluate progressively pruned and recovered D3D-8B checkpoints under asynchronous inference (Figure~\ref{fig:layer_pruning_ablation}, left).
Pruning from 36 to 31 layers reduces depth by $13.9\%$ while retaining over $99\%$ of mean $\tau$ across MATH-500, HumanEval, and MT-Bench.
The 31-layer checkpoint achieves the highest mean decoding speedup among the tested depths, at $4.69\times$.
The small acceptance loss indicates that part of the original depth is redundant for drafting after recovery.
Reducing drafting work makes it easier to overlap drafting with verification, while the retained acceptance keeps the average number of output tokens per round nearly unchanged.
Below 31 layers, acceptance length and speedup both decline, indicating that further computational savings no longer compensate for the loss in draft quality.
We therefore select the final recovered checkpoint by validation decoding speedup.

\paragraph{Pretrained initialization.}
Holding the drafter architecture and training budget fixed, we compare Qwen3-8B weights with random initialization to isolate the contribution of pretraining (Figure~\ref{fig:layer_pruning_ablation}, right).
At the final measured training step, pretrained initialization yields $\tau\approx9.8$ on GSM8K, compared with $\tau\approx4.5$ for random initialization; the advantage emerges early and persists throughout training.
A pretrained model already predicts plausible continuations from a token prefix, allowing adaptation to focus on target alignment and parallel block prediction.
A randomly initialized drafter must also learn this prediction capability within the same training budget.

\begin{figure}[!htb]
  \centering
  \begin{minipage}[t]{0.48\linewidth}
    \centering
    \includegraphics[width=\linewidth]{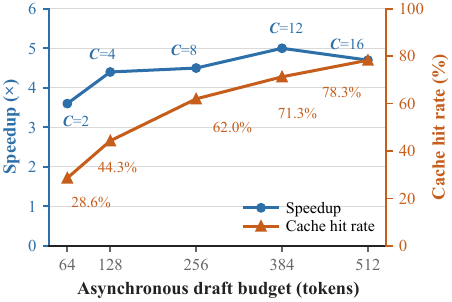}
  \end{minipage}
  \hspace{0.01\linewidth}%
  \begin{minipage}[t]{0.48\linewidth}
    \centering
    \includegraphics[width=\linewidth]{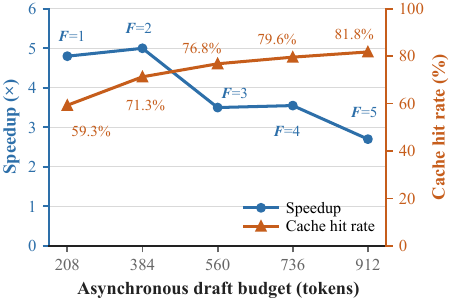}
  \end{minipage}
\caption{Decoding speedup and cache hit rate versus asynchronous draft budget on AIME25, using the 31-layer D3D-8B drafter with Qwen3-8B and $B=16$. Left: varying position coverage $C$ at $F=2$. Right: varying token fan-out $F$ at $C=12$.}
  \label{fig:async_budget}
\end{figure}

\paragraph{Asynchronous draft budget.}
Two sweeps examine the tradeoff between speculative coverage and drafting cost: varying position coverage $C$ at fixed $F=2$, and varying token fan-out $F$ at fixed $C=12$ (Figure~\ref{fig:async_budget}). We use the 31-layer D3D-8B drafter on AIME25.
Both sweeps achieve their highest speedup of approximately $5.0\times$ at $C=12$, $F=2$, and a 384-token asynchronous draft budget. Increasing $C$ from 12 to 16 at $F=2$ raises the hit rate from $71.3\%$ to $78.3\%$ but reduces speedup. At $C=12$, increasing $F$ from 2 to 5 raises the hit rate to $81.8\%$ and the budget to 912 tokens, with lower speedup than at $F=2$. Cache hits avoid a new drafting pass after verification, reducing the delay before the next round. Increasing coverage adds computation to every packed drafting pass, including branches that verification never selects. We therefore select the budget by decoding speedup to account for cache reuse and the cost of preparing successor blocks.

%% file: sections/conclusion.tex
\section{Conclusion}

We presented D3D, which provides the first empirical demonstration of asynchronous lossless speculative decoding with a standalone diffusion drafter.
Target-aligned training combines blockwise denoising with auxiliary autoregressive supervision to adapt a pretrained model for single-pass block prediction without target hidden states.
Acceptance-guided progressive pruning with recovery training reduces drafting latency with little loss in acceptance length.
Across seven benchmarks, D3D achieves average decoding speedups of $5.00\times$ and $5.71\times$ over autoregressive decoding with Qwen3-8B and Qwen3-32B as targets, respectively.
These average speedups exceed those of both SSD and asynchronously deployed PARD-2 in our evaluation.
With Qwen3-8B as the target, pruning D3D-8B from 36 to 31 layers retains over $99\%$ of the unpruned drafter's mean acceptance length across three benchmarks. These results show that target-aligned training and acceptance-guided pruning make standalone diffusion drafting effective for asynchronous lossless speculative decoding.

\section*{Limitations}
We evaluate D3D, SSD, and PARD-2 with one GPU for the target and another for the drafter.
The drafter adds model memory and speculative computation.
We have not compared throughput per GPU against alternative uses of the same two-GPU budget, such as target inference using tensor parallelism.
Cache misses require synchronous drafting before verification can resume.
Our implementation uses Transformers; evaluating D3D in production frameworks such as SGLang or vLLM with their batching and scheduling optimizations remains future work.

%% file: sections/appendix.tex
\appendix

\section{Teacher Distribution Adjustment}
\label{app:teacher_distribution_adjustment}

At each masked prediction position, let $q^{(0)}$ denote the fixed teacher's token distribution and $y$ the supervised token from the target-generated response.
The teacher's probability margin for $y$ is
\begin{equation}
m=q^{(0)}(y)-\max_{v\neq y}q^{(0)}(v).
\end{equation}
The mixture in Section~\ref{sec:acceptance_guided_layer_pruning} changes this margin to $m+\eta(1-m)$.
For a required margin $\delta\in(0,1)$, the smallest mixing coefficient that satisfies $m+\eta(1-m)\geq\delta$ is
\begin{equation}
\eta=\max\!\left(0,\frac{\delta-m}{1-m}\right).
\end{equation}
When $m\geq\delta$, $\eta=0$ leaves the teacher distribution unchanged.
Otherwise, the adjustment gives $y$ a margin of $\delta$ while retaining the largest possible teacher contribution $1-\eta$.
The mixture scales all non-target probabilities by $1-\eta$, preserving their ratios.
We use $\delta=0.01$ in our experiments.

\section{Additional Ablations}
\label{app:additional_ablations}

\subsection{Training Dynamics}
\label{app:training_dynamics}
\begin{figure}[!htb]
  \centering
  \includegraphics[width=0.66\linewidth,trim=10 10 15 25,clip]{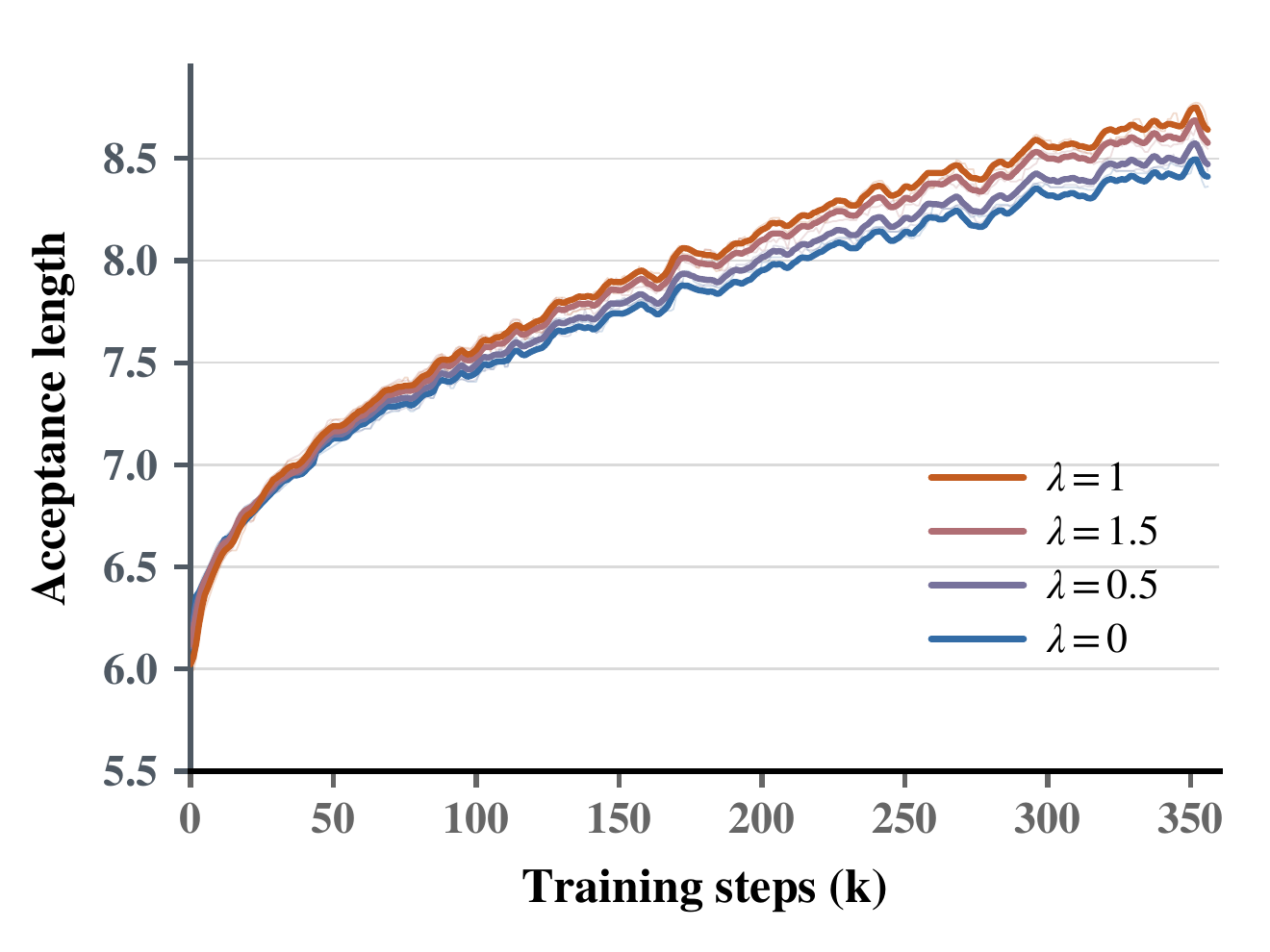}
\caption{GSM8K acceptance length $\tau$ during D3D-1.7B training with different auxiliary autoregressive loss weights $\lambda$. The target is Qwen3-8B and the block size is $B=16$.}
  \label{fig:ablation_loss_weight}
\end{figure}

We examine whether auxiliary AR supervision provides a sustained benefit by tracking GSM8K acceptance length during one-epoch D3D-1.7B training (Figure~\ref{fig:ablation_loss_weight}).
After the initial training phase, nonzero AR weights yield higher acceptance than diffusion-only training ($\lambda=0$). Among the tested weights, $\lambda=1$ generally gives the highest acceptance; increasing it to $1.5$ provides no further benefit. The sustained advantage supports maintaining next-token supervision during diffusion adaptation, where it continues to constrain predictions from unmasked prefixes.

\subsection{Block Size}
\label{app:block_size_ablation}
To examine how block size affects acceptance and decoding speedup, we train separate D3D-8B drafters for one epoch at $B\in\{8,16,32\}$.
We evaluate each drafter at its training block size under asynchronous inference with Qwen3-8B (Table~\ref{tab:block_size_ablation}).
Increasing $B$ from 8 to 16 improves speedup on MATH-500 and HumanEval, while MT-Bench is slightly faster at $B=8$.
Increasing $B$ further to 32 raises acceptance length but reduces speedup on all three tasks; on MATH-500, $\tau$ increases from 9.89 to 12.39 while speedup falls from $5.55\times$ to $3.16\times$.
Larger blocks permit longer accepted prefixes but also add candidate positions for drafting and verification, including tokens that verification may discard.
We use $B=16$ as the default because it achieves the highest mean speedup across the three tasks.

\begin{table}[!htb]
  \centering
  \caption{Decoding speedup and acceptance length $\tau$ versus block size; D3D-8B, Qwen3-8B target.}
  \label{tab:block_size_ablation}
  \small
  \setlength{\tabcolsep}{4pt}
  \begin{tabular}{@{}c cc cc cc@{}}
    \toprule
    \multirow{2}{*}{$B$} & \multicolumn{2}{c}{MATH-500} & \multicolumn{2}{c}{HumanEval} & \multicolumn{2}{c}{MT-Bench} \\
    \cmidrule(lr){2-3}\cmidrule(lr){4-5}\cmidrule(lr){6-7}
    & Speedup & $\tau$ & Speedup & $\tau$ & Speedup & $\tau$ \\
    \midrule
    8 & $4.58\times$ & 6.26 & $4.00\times$ & 5.52 & \textbf{$2.55\times$} & 4.38 \\
    16 & \textbf{$5.55\times$} & 9.89 & \textbf{$4.47\times$} & 7.85 & $2.47\times$ & 5.72 \\
    32 & $3.16\times$ & \textbf{12.39} & $2.56\times$ & \textbf{8.70} & $1.42\times$ & \textbf{6.33} \\
    \bottomrule
  \end{tabular}
\end{table}

\subsection{Number of Training Anchors}
\label{app:anchor_count_ablation}
The number of sampled anchors controls how many diffusion blocks each response contributes to training.
We test whether denser supervision improves acceptance by training D3D-1.7B for one epoch with $K\in\{16,64,128,256\}$ anchors per response, using Qwen3-8B as the target and $B=16$ (Table~\ref{tab:anchor_count_ablation}).
Increasing $K$ from 16 to 64 improves $\tau$ by 0.25, 0.21, and 0.04 tokens on MATH-500, HumanEval, and MT-Bench, respectively.
Increasing $K$ from 64 to 256 adds only 0.03, 0.03, and 0.06 tokens.
Given that target-generated responses average fewer than 2,048 tokens, the limited gains suggest that $K=128$ at $B=16$ already provides sufficient supervision across response positions.
Doubling $K$ adds more overlapping blocks and training computation, but improves acceptance length by at most 0.02 tokens on each task.
We therefore use $K=128$.

\begin{table}[!htb]
  \centering
  \caption{Average acceptance length $\tau$ versus the number of sampled training anchors; D3D-1.7B, Qwen3-8B target, $B=16$.}
  \label{tab:anchor_count_ablation}
  \small
  \setlength{\tabcolsep}{5pt}
  \begin{tabular}{@{}cccc@{}}
    \toprule
    $K$ & MATH-500 & HumanEval & MT-Bench \\
    \midrule
    16 & 7.89 & 6.00 & 4.41 \\
    64 & 8.14 & 6.21 & 4.45 \\
    128 & 8.15 & 6.24 & 4.50 \\
    256 & 8.17 & 6.24 & 4.51 \\
    \bottomrule
  \end{tabular}
\end{table}

\section{Selective Prefilling with D3D}
\label{app:selective_prefilling}

Our main results use a trained standalone diffusion drafter for full-prompt
asynchronous speculative decoding. For each drafter size, we use a single trained checkpoint
for both prompt selection and sequential speculative decoding. Prefilling fewer tokens reduces target computation and
can further improve end-to-end latency. Because the target conditions on a
reduced prompt, we report task scores alongside latency speedups to quantify
the quality--latency trade-off. Decoding remains target-verified for the
retained prompt. SpecPrefill~\citep{liu2025specprefill} suggests reusing its
token selector for speculative decoding but provides no experimental results
for the combination. To our knowledge, this is the first long-context
evaluation in which a standalone diffusion drafter performs both prompt
selection and speculative drafting.

\begin{figure}[!ht]
  \centering
  \includegraphics[width=\linewidth]{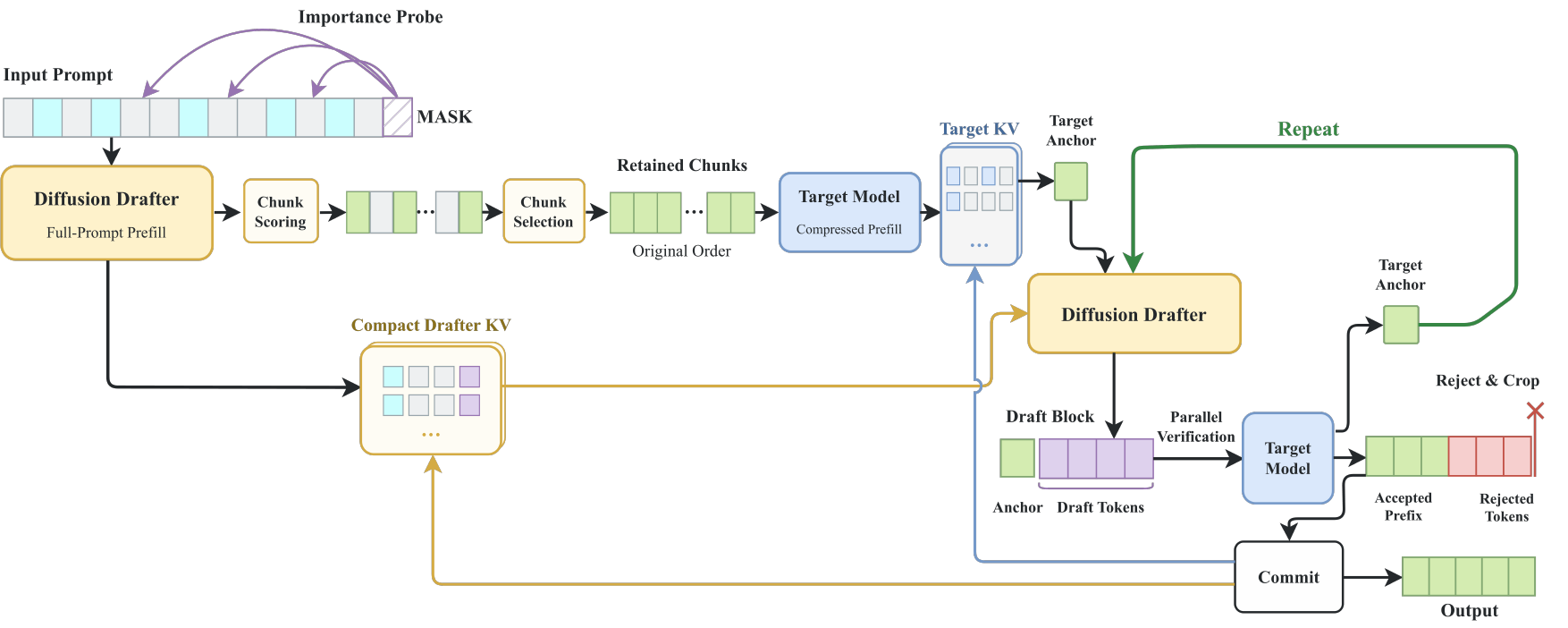}
\caption{Selective prefilling and sequential speculative decoding with D3D. The same standalone drafter scores prompt chunks for target prefilling and generates candidate blocks for verification. Retained chunks preserve their original order, and the drafter and target maintain separate KV caches.}
  \label{fig:selective_prefill_workflow}
\end{figure}

\paragraph{Inference setting.}
Figure~\ref{fig:selective_prefill_workflow} illustrates the inference workflow.
For a prompt $x=(x_1,\ldots,x_L)$, we run the standalone drafter on the full
prompt and score prompt chunks with a one-MASK importance probe. Given a
target retention fraction $\rho\in\{0.1,0.2,0.3\}$, we select the
highest-scoring chunks to retain approximately $\rho L$ tokens, preserving
their original order to obtain $x_\rho$.
The target model prefills $x_\rho$; the same drafter then prepares
blocks of size 16 (one target anchor and 15 drafted tokens) for sequential,
target-verified speculative decoding. The drafter maintains its own prefix
cache, separate from the target cache built on $x_\rho$.
Greedy verification follows the target model conditioned on $x_\rho$;
selecting $x_\rho$ can change the output relative to conditioning on the
original prompt $x$.

\paragraph{Evaluation.}
We evaluate Qwen3-32B on 18 datasets across five task categories from the
original LongBench~\citep{bai2024longbench}. We compare full-prompt
sequential D3D drafting with selective-prefill variants retaining 10\%, 20\%,
or 30\% of the prompt.
Tables~\ref{tab:prefill_17b} and~\ref{tab:prefill_8b} report the LongBench
score for each category and the speedup in time to first token (TTFT), decoding,
and end-to-end (E2E) latency relative to full-prompt AR decoding in the same
category. ``Full'' denotes the complete prompt; percentages denote the
target retention fraction. Higher is better in every column. Category scores use the
corresponding LongBench task metrics and should be compared within a category.

\begin{table}[!ht]
  \centering
  \caption{LongBench results with a 1.7B D3D drafter and Qwen3-32B target model.}
  \label{tab:prefill_17b}
  \small
  \setlength{\tabcolsep}{4pt}
  \renewcommand{\arraystretch}{0.98}
  \begin{tabular*}{\linewidth}{@{\extracolsep{\fill}}llrrrr@{}}
    \toprule
    Task & Setting & Score & TTFT & Decode & E2E \\
    \midrule
    Single-doc QA & Full + D3D & 47.91 & 0.84$\times$ & 3.09$\times$ & 1.46$\times$ \\
    & 30\% + D3D & 44.23 & 2.09$\times$ & 3.81$\times$ & 2.83$\times$ \\
    & 20\% + D3D & 42.46 & 2.57$\times$ & 3.95$\times$ & 3.22$\times$ \\
    & 10\% + D3D & 41.27 & 2.91$\times$ & 4.25$\times$ & 3.56$\times$ \\
    \midrule
    Multi-doc QA & Full + D3D & 45.98 & 0.86$\times$ & 2.24$\times$ & 1.51$\times$ \\
    & 30\% + D3D & 44.32 & 2.01$\times$ & 2.07$\times$ & 2.05$\times$ \\
    & 20\% + D3D & 41.27 & 2.71$\times$ & 2.47$\times$ & 2.54$\times$ \\
    & 10\% + D3D & 40.68 & 2.84$\times$ & 2.28$\times$ & 2.42$\times$ \\
    \midrule
    Summarization & Full + D3D & 24.07 & 0.81$\times$ & 1.82$\times$ & 1.76$\times$ \\
    & 30\% + D3D & 22.71 & 1.74$\times$ & 2.24$\times$ & 2.23$\times$ \\
    & 20\% + D3D & 21.88 & 2.55$\times$ & 2.62$\times$ & 2.62$\times$ \\
    & 10\% + D3D & 21.17 & 3.54$\times$ & 2.98$\times$ & 3.00$\times$ \\
    \midrule
    Few-shot learning & Full + D3D & 57.07 & 0.85$\times$ & 2.26$\times$ & 1.79$\times$ \\
    & 30\% + D3D & 57.72 & 1.30$\times$ & 1.64$\times$ & 1.57$\times$ \\
    & 20\% + D3D & 57.79 & 2.41$\times$ & 2.38$\times$ & 2.39$\times$ \\
    & 10\% + D3D & 55.84 & 3.25$\times$ & 2.43$\times$ & 2.53$\times$ \\
    \midrule
    Code completion & Full + D3D & 14.73 & 0.90$\times$ & 1.69$\times$ & 1.52$\times$ \\
    & 30\% + D3D & 13.62 & 2.05$\times$ & 2.15$\times$ & 2.14$\times$ \\
    & 20\% + D3D & 13.19 & 2.43$\times$ & 2.14$\times$ & 2.16$\times$ \\
    & 10\% + D3D & 13.28 & 2.71$\times$ & 2.01$\times$ & 2.07$\times$ \\
    \bottomrule
  \end{tabular*}
\end{table}

\begin{table}[!ht]
  \centering
  \caption{LongBench results with an 8B D3D drafter and Qwen3-32B target model.}
  \label{tab:prefill_8b}
  \small
  \setlength{\tabcolsep}{4pt}
  \renewcommand{\arraystretch}{0.98}
  \begin{tabular*}{\linewidth}{@{\extracolsep{\fill}}llrrrr@{}}
    \toprule
    Task & Setting & Score & TTFT & Decode & E2E \\
    \midrule
    Single-doc QA & Full + D3D & 47.91 & 0.72$\times$ & 4.16$\times$ & 1.40$\times$ \\
    & 30\% + D3D & 47.42 & 1.50$\times$ & 5.49$\times$ & 2.62$\times$ \\
    & 20\% + D3D & 46.41 & 1.70$\times$ & 5.96$\times$ & 2.93$\times$ \\
    & 10\% + D3D & 47.05 & 1.93$\times$ & 6.73$\times$ & 3.32$\times$ \\
    \midrule
    Multi-doc QA & Full + D3D & 45.97 & 0.74$\times$ & 2.82$\times$ & 1.54$\times$ \\
    & 30\% + D3D & 46.25 & 1.54$\times$ & 3.15$\times$ & 2.40$\times$ \\
    & 20\% + D3D & 46.95 & 1.80$\times$ & 3.18$\times$ & 2.59$\times$ \\
    & 10\% + D3D & 45.75 & 2.10$\times$ & 3.24$\times$ & 2.79$\times$ \\
    \midrule
    Summarization & Full + D3D & 24.06 & 0.72$\times$ & 2.29$\times$ & 2.16$\times$ \\
    & 30\% + D3D & 23.06 & 1.59$\times$ & 3.48$\times$ & 3.37$\times$ \\
    & 20\% + D3D & 22.16 & 1.78$\times$ & 3.79$\times$ & 3.68$\times$ \\
    & 10\% + D3D & 21.69 & 2.14$\times$ & 4.29$\times$ & 4.18$\times$ \\
    \midrule
    Few-shot learning & Full + D3D & 57.06 & 0.73$\times$ & 2.99$\times$ & 2.04$\times$ \\
    & 30\% + D3D & 56.19 & 1.60$\times$ & 3.61$\times$ & 3.03$\times$ \\
    & 20\% + D3D & 55.86 & 1.87$\times$ & 3.72$\times$ & 3.23$\times$ \\
    & 10\% + D3D & 55.43 & 2.15$\times$ & 3.46$\times$ & 3.16$\times$ \\
    \midrule
    Code completion & Full + D3D & 14.73 & 0.68$\times$ & 2.21$\times$ & 1.76$\times$ \\
    & 30\% + D3D & 14.16 & 1.54$\times$ & 2.77$\times$ & 2.54$\times$ \\
    & 20\% + D3D & 13.38 & 1.79$\times$ & 2.84$\times$ & 2.66$\times$ \\
    & 10\% + D3D & 13.24 & 1.94$\times$ & 2.74$\times$ & 2.62$\times$ \\
    \bottomrule
  \end{tabular*}
\end{table}

\FloatBarrier

With Full + D3D, the target still prefills the entire prompt, while
sequential execution adds drafter prefilling, first-block drafting, and
verification before the first token. TTFT speedups therefore fall below
$1\times$ despite faster decoding.

\noindent\begin{minipage}{\linewidth}
Selective prefilling adds substantial end-to-end gains to D3D's decoding
acceleration. With the 8B drafter, 30\% retention improves E2E speedup
over Full + D3D in all five LongBench categories, spanning document QA,
summarization, few-shot learning, and code completion, while keeping scores
close to full-prompt results. The gains persist at lower retention: on
single-document QA, 10\% raises E2E speedup from 1.40$\times$ to
3.32$\times$ with a score of 47.05 versus 47.91 for Full + D3D; at 20\%,
multi-document QA scores 46.95 at 2.59$\times$ E2E speedup. The same
trained standalone drafter thus accelerates target prefilling and decoding
across tasks and retention levels.
\end{minipage}

\medskip
\noindent\begin{minipage}{\linewidth}
\paragraph{Drafter size.}
A larger drafter improves QA under aggressive selective prefilling.
At 10\% retention, the 8B drafter scores 47.05 on single-document QA
versus 41.27 for the 1.7B drafter; both score 47.91 with the full prompt.
The gap appears when the drafter selects the target's context, indicating
that larger capacity helps preserve the information needed for accurate
answers with a much shorter target prompt.
\end{minipage}

\paragraph{Prompt retention.}
A smaller target prompt can add to the E2E gain from speculative drafting.
On multi-document QA with the 8B drafter, retaining 20\% of the prompt
scores 46.95 versus 45.97 with Full + D3D, and increases E2E speedup
from 1.54$\times$ to 2.59$\times$. Selection reduces the target's prefill
work while keeping the context needed for answering; the same drafter
continues to accelerate decoding.

\paragraph{Speedup across tasks.}
The balance between prefilling and decoding determines how much decoding
acceleration translates into E2E speedup. With the 8B drafter at 30\%
retention, summarization reaches
3.37$\times$ E2E speedup, close to its 3.48$\times$ decoding speedup.
Single-document QA reaches 2.62$\times$ E2E despite 5.49$\times$ faster
decoding. Summarization spends more of its runtime generating the response,
while prefilling accounts for a larger share in QA. Accelerating both
stages yields substantial E2E gains across these different task profiles.